# Enhancing Extractive Text Summarization with Topic-Aware Graph Neural Networks

**Peng Cui, Le Hu,** and **Yuanchao Liu**[*]
School of Computer Science and Technology, Harbin Institute of Technology
`{pcui, lhu, lyc}@insun.hit.edu.cn`

## Abstract

Text summarization aims to compress a textual document to a short summary while keeping salient information. Extractive approaches are widely used in text summarization because of their fluency and efficiency. However, most of existing extractive models hardly capture inter-sentence relationships, particularly in long documents. They also often ignore the effect of topical information on capturing important contents. To address these issues, this paper proposes a graph neural network (GNN)-based extractive summarization model, enabling to capture inter-sentence relationships efficiently via graph-structured document representation. Moreover, our model integrates a joint neural topic model (NTM) to discover latent topics, which can provide document-level features for sentence selection. The experimental results demonstrate that our model not only substantially achieves state-of-the-art results on CNN/DM and NYT datasets but also considerably outperforms existing approaches on scientific paper datasets consisting of much longer documents, indicating its better robustness in document genres and lengths. Further discussions show that topical information can help the model preselect salient contents from an entire document, which interprets its effectiveness in long document summarization.

## 1 Introduction

Text summarization is an important task in natural language processing, which can help people rapidly acquire important information from a large sum of documents. Previous summarization approaches can be mainly classified into two categories, which are abstractive and extractive. Neural-based abstractive models usually use a seq2seq framework (Sutskever et al., 2014) to generate a word-by-word summary after encoding a full document. By contrast, extractive models directly select important sentences from the original document and then aggregate them into a summary. Abstractive models are generally more flexible but may produce disfluent or ungrammatical summary texts (Liu and Lapata, 2019b), whereas extractive models have advantages in factuality and efficiency (Cao et al., 2018).

Despite their success, modeling long-range inter-sentence relationships for summarization remains a challenge (Xu et al., 2019b). Hierarchical networks are usually applied to tackle this problem by modeling a document as a sequence of sequences (Cohan et al., 2018; Zhang et al., 2019). However, empirical observations (Liu and Lapata, 2019a) showed that the use of such a paradigm to model inter-sentence relationships does not provide much performance gain for summarization. Hierarchical approaches are also slow to train and tend to overfit (Xiao and Carenini, 2019). Most recently, Graph Neural Networks (GNNs) are widely explored to model cross-sentence relationships for summarization task. The critical step of this framework is to build an effective document graph. Several studies (Xu et al., 2019a; Yasunaga et al., 2017) built document graphs based on discourse analysis. However, this approach depends on external tools and may lead to other problems, such as semantically fragmented output (Liu et al., 2019). Wang and Liu (2020) built a word–sentence document graph based on word appearance, but such statistical graph-building approach hardly captures semantic-level relationships. Therefore, how to model a document as a graph for summarization effectively remains an open question.

---

[*] Corresponding author

Another critical point of summarization is modeling global information, which plays a key role in sentence selection (Xiao and Carenini, 2019). Pre-trained language models can considerably boost the performance of summarization (Liu and Lapata, 2019a; Zhang et al., 2019) since they effectively capture context features. However, they are poor at modeling document-level information, particularly for long documents, because most of them are designed for sentences or a short paragraph (Xu et al., 2019b).

To tackle the abovementioned weaknesses, this paper proposes a novel graph-based extractive summarization model. First, we encode an entire document with a pre-trained BERT (Devlin et al., 2019) to learn contextual sentence representations, and discover latent topics with a joint neural topic model (NTM; Miao et al., 2017; Srivastava and Sutton, 2017). Second, we build a heterogeneous document graph consisting of sentence and topic nodes, and simultaneously update their representations with a modified graph attention network (GAT; Veličković et al., 2017). Third, the representations of sentence nodes are extracted to compute the final labels. Intuitively, our topic–sentence document graph has the following advantages: 1) During the graph propagation, sentence representations can be enriched by topical information, which can be considered as a kind of document-level feature and help our model distil important contents from an entire document. 2) Topic nodes can act as intermediary to bridge long-distance sentences; hence, our model can efficiently capture inter-sentence relationships. We evaluate our model on four standard datasets, including news articles and scientific papers. The experimental results show its effectiveness and superiority. To summarize, our contributions are threefold.

- We conduct a quantitative exploration on the effect of latent topics on document summarization and provide an intuitive understanding of how topical information help summarize documents.

- We propose a novel graph-based neural extractive summarization model, which innovatively incorporates latent topics into graph propagation via a joint neural topic model. To the best of our knowledge, we are the first to propose applying NTM to the extractive text summarization task.

- The experimental results demonstrate that our proposed model not only achieves competitive results compared with state-of-the-art extractive models on news datasets but also considerably outperforms existing approaches on scientific paper datasets consisting of much longer documents, indicating its better robustness in document genres and lengths.

## 2 Related Work

**Neural Extractive Summarization** Neural networks have achieved remarkable results in extractive summarization. Existing works mainly regard extractive summarization as a sequence labeling task (Nallapati et al., 2017; Zhang et al., 2018; Dong et al., 2018) or sentence ranking task (Narayan et al., 2018). Pre-trained language models have provided substantial performance gain for summarization (Liu and Lapata, 2019a; Zhang et al., 2019; Xu et al., 2019). In the current work, we further model inter-sentence relationships with a graph encoder and enrich sentence representations with topical information after a BERT encoder.

**Graph-based Summarization** Early works, such as TextRank (Mihalcea and Tarau, 2004) and LexRank (Erkan and Radev, 2004), built document graphs on the basis of inter-sentence similarity and extracted summary sentences in an unsupervised manner. Recently, the application of GNNs to document summarization has attracted considerable interests (Yasunaga et al., 2017; Xu et al., 2019b; Fernandes et al., 2018; Wang and Liu et al., 2020). Existing GNN-based summarization models build document graphs on the basis of only words or sentences. On the contrary, we explore the effects of high-level semantic units, i.e., latent topics.

**Topic Modeling for Summarization** Topic modeling is a powerful approach to learning document features. However, it has been rarely applied to document summarization. Wei et al. (2012) proposed to build a document graph consisting of words, sentences, and topic nodes and learn the graph with Markov chain. Zheng et al. (2019) proposed to summarize multiple documents by mining cross-document subtopics. Narayan et al. (2018) recommended enriching word representation with topical information. Unlike them, we discover latent topics with a neural topic model together with summarization. To the best of our knowledge, NTM had never been applied to extractive summarization task.

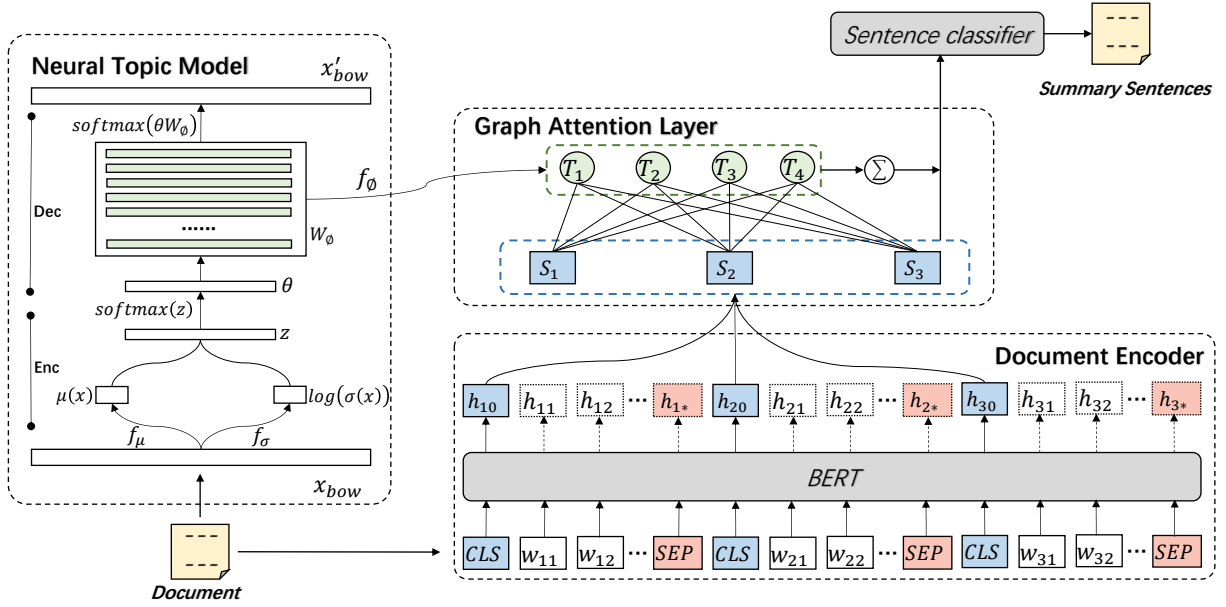

Figure 1. Overall architecture of our model (**Topic-GraphSum**). In the graph attention layer (top right), the **square nodes** denote the sentence representations output from the document encoder (bottom right), and the **circular nodes** denote the topic representations learned by NTM (left).

## 3 Model

This section describes our model, namely, topic-aware graph neural network for document summarization (**Topic-GraphSum**). Figure 1 presents the overview architecture. Given an arbitrary document $D = \{s_1, s_2, ..., s_N\}$ that consists of $N$ sentences, the objective of our model is to learn a sequence of binary labels $\{y_1, y_2, ..., y_N\}$, where $y_i \epsilon \{0, 1\}$ represents whether the $i$-th sentence should be included in summary. Our model generally consists of three parts, which are the 1) **document encoder**, 2) **neural topic model**, and 3) **graph attention layer**. Given the input document, the document encoder learns contextual representations of each sentence with a pre-trained BERT. The NTM aims to learn the document topic distribution and a group of topic representations. The graph attention layer builds a heterogeneous document graph with topics and sentences and then simultaneously update their node representations. After graph encoding, sentence representations are further combined with topics and then sent to a sentence classifier to compute the final labels. We elucidate each part below.

### 3.1 Document Encoder

BERT is a bidirectional transformer encoder pre-trained with a large corpus. Similar to previous works (Xu et al., 2019b; Liu and Lapata, 2019a), we employ a modified version of BERT to generates local context-aware hidden representations of sentences. Specifically, we insert $<CLS>$ and $<SEP>$ tokens at the beginning and end of each sentence, respectively. Then, we put all tokens into BERT layer and learn their hidden states.

$$\{h_{1,0}, h_{1,1}, ..., h_{N,0}, ..., h_{N,*}\} = BERT(w_{1,0}, w_{1,1}, ..., w_{N,0}, ..., w_{N,*}), \qquad (1)$$

where $w_{i,j}$ represents the $j$-th word of the $i$-th sentence. $w_{i,0}$ and $w_{i,*}$ represent the $<CLS>$ and $<SEP>$ tokens of the $i$-th sentence, and $h_{i,j}$ represents the hidden state of the corresponding token.

After the BERT encoding, we regard the hidden states of $<CLS>$: $\boldsymbol{H_B} = \{h_{1,0}, ..., h_{N,0}\}$ as the corresponding sentence contextual representations, which will be further enriched by topic information.

### 3.2 Neural Topic Model

NTM is based on the Variational Autoencoder (VAE; Kingma and Welling, 2013) framework. It learns the latent topic via an encoding–decoding process. Let $x_{bow} \in \mathbb{R}^{|V|}$ be the bag-of-words representation of a given document, where $V$ is the vocabulary. In the encoder, we have $\mu = f_\mu(x), log\sigma = f_\sigma(x)$, where $\mu$ and $\sigma$ are the prior parameters for parameterizing topic distribution in decoder networks. Functions $f_\mu$ and $f_\sigma$ are linear transformations with ReLU activation.

The decoder can be regarded as a three-step document generation process. First, we employ Gaussian softmax (Miao et al., 2017) to draw topic distribution, i.e., $z \sim \mathcal{N}(\mu, \sigma^2), \theta = softmax(z)$, where $z$ is the latent topic variable, $\theta \in \mathbb{R}^K$ is the topic distribution, and $K$ is the predefined topic number. Second, we learn the probability of predicted words $p_w \in \mathbb{R}^{|V|}$ throughout $p_w = softmax(W_\emptyset \theta)$. $W_\emptyset \in \mathbb{R}^{|V| \times K}$ is analogous to the topic–word distribution matrix in LDA-style topic models, and $W_\emptyset^{(i,j)}$ represents the relevance between the $i$-th word and $j$-th topic. Finally, we draw each word from $p_w$ to reconstruct input $x_{bow}$. We leave out the details and refer the readers to Miao et al. (2017).

Considering the intermediate parameters $W_\emptyset$ and $\theta$ have encoded topical information, we further use them to build topic representations as follows:

$$H_T = f_\emptyset(W_\emptyset^T), \quad (2)$$

$$T_d = \sum_{1 \leq i \leq K} \theta^{(i)} H_T^{(i)}, \quad (3)$$

where $H_T \in \mathbb{R}^{K \times d_t}$ represents a group of topic representations with a predefined dimension of $d_t$, and $f_\emptyset$ is a linear transformation with ReLU activation. $T_d \in \mathbb{R}^{d_t}$ is the weighted sum of each topic representation, which can be regarded as the overall topic representation of document.

$H_T$ and $T_d$ are used in the graph attention layer to enrich sentence representation. Other summarization approaches (Zheng et al., 2019; Narayan et al., 2018) with topical information learn topic as a fixed feature from an external model. In comparison with them, the latent topic of our model is learned via a neural approach and can be dynamically updated with entire networks.

### 3.3 Graph Attention Layer

**Graph Building** Let $G = \{V, E\}$ represent an arbitrary graph, where $V$ represents the node set and $E$ represents the edge set. Formally, our undirected graph can be defined as. $V = V_S \cup V_T$; $E = \{e_{1,1}, \dots, e_{N,K}\}$, where $V_S = \{s_1, s_2, \dots, s_N\}$ stands for $N$ sentence nodes and $V_T = \{T_1, T_2 \dots, T_K\}$ stands for $K$ topic nodes. $e_{i,j}$ represents the edge between the $i$-th sentence and $j$-th topic, indicating that our document graph is bipartite.

**Graph Propagation** We initialize the vectors of sentence nodes and that of topic nodes with $H^B$ learned from the document encoder and $H^T$ learned from NTM (Eq. 2), respectively. Then, we update node representations with graph attention network, which can be denoted as:

$$z_{i,j} = LeaklyReLU(W_a[W_e h_i; W_e h_j]), \quad (4)$$

$$\alpha_{i,j} = \frac{exp(z_{i,j})}{\sum_{l \in \mathcal{N}_i} exp(z_{i,l})}, \quad (5)$$

$$u_i = \sum_{j \in \mathcal{N}_i} tanh(\alpha_{i,j} W_c h_j), \quad (6)$$

$$h_i' = ||_{m=1}^M \sum_{j \in \mathcal{N}_i} tanh(\alpha_{i,j}^m W_c^m h_j), \quad (7)$$

where $h_i$ is the $i$-th node representation, and $\mathcal{N}_i$ represents its neighbor nodes. $\|*$ represents multi-heads concatenation. $W_a$, $W_e$, and $W_c$ are model trainable parameters.

The vanilla GAT is designed for homogeneous graphs. However, our document graph is heterogeneous because the sentence and topic should be considered different semantic units; hence, we need to make some adaptation. Inspired by Hu et al. (2019), we consider a convenient approach to project the topic and sentence representations into an implicit common space, in which we calculate the attention weight. Let $h_i$ be the $i$-th sentence node and $h_j$ be the $j$-th topic node. We modify Eq. 4 by replacing shared matrix $W_e$ with different projection functions, as shown as follows:

$$z_{i,j} = LeaklyReLU(W_a[f_S(h_i); f_T(h_j)]), \tag{8}$$

where $f_S$ and $f_T$ are the nonlinear transformation functions to project sentence and topic nodes to a common vector space, respectively.

The graph attention layer can build semantic relationships between sentences and topics. For example, during graph propagation, sentences can enrich their representation with topical information, which can be regarded as a global feature. Topics can capture their related sentences and distil salient contents from an entire document by their different topical relevance. Meanwhile, topic nodes can act as intermediary to help build inter-sentence relationships because they are high-level semantic units across sentences.

After graph encoding, we obtain topic-sensitive sentence representations. We concatenate them with overall topic representation $T_d$ (Eq. 3) to further capture their topical relevance to the document. Then, we choose a single feed-forward layer as the sentence classifier[1] to predict the final labels, i.e., $\hat{y}_i = \sigma([h_i: T_d])$, where $\sigma(*)$ is the sigmoid function.

## 3.4 Joint Training

We jointly train NTM and sentence classifier. For the NTM, the objective function is defined as the negative evidence lower bound, as shown as follows:

$$\mathcal{L}_{NTM} = D_{KL}(p(z)||q(z|x)) - \mathbb{E}_{q(z|x)}[p(x|z)], \tag{9}$$

where the first term indicates the Kullback–Leibler divergence loss, and the second term indicates the reconstruction loss. $q(z|x)$ and $p(x|z)$ represent the encoder and decoder networks, respectively.

The binary cross-entropy loss of the sentence classifier is expressed as:

$$\mathcal{L}_{SC} = \sum_{i=1}^{n}(y_i \log(\hat{y}_i) + (1 - y_i)\log(1 - \hat{y}_i)), \tag{10}$$

The final loss of our model is the linear combination of two parts of loss with hyperparameter $\lambda$ to balance their weights, i.e.,

$$\mathcal{L} = \mathcal{L}_{SC} + \lambda \mathcal{L}_{NTM}. \tag{11}$$

## 4 Experimental Setup

### 4.1 Datasets

We conduct experiments on four datasets, including two document types, which are news article and scientific paper. The summarization of news articles has been widely explored, but that of much longer scientific papers is more challenging since accurately encoding long texts for summarization is a known challenge (Vaswani et al., 2017; Frermann and Klementiev, 2019). Therefore, we conduct experiments on scientific paper datasets to verify the generalization capability of our model for long documents. The detailed statistics of four datasets is summarized in Table 1.

| Datasets | Source | # Docs | | | # Avg. Tokens | |
|---|---|---|---|---|---|---|
| | | Train | Val | Test | Doc. | Sum. |
| CNN | News | 90,266 | 1,220 | 1,093 | 761 | 46 |
| Daily Mail | News | 196,961 | 12,148 | 10,397 | 653 | 55 |
| NYT | News | 96,834 | 4,000 | 3,452 | 800 | 46 |
| arXiv | Scientific Paper | 203,037 | 6,436 | 6,440 | 4,938 | 220 |
| PubMed | Scientific Paper | 119,924 | 6,633 | 6,658 | 3,016 | 203 |

Table 1: Statistics of four datasets: split size, average tokens of document and summary.

---

[1] We also tried adding more advanced classifiers (e.g., CNN and RNN) on top of GAT layer. However, the performance shows no substantial gain, indicating that our model has already learned sufficient features.

**CNN/DailyMail** (Hermann et al., 2015) is the most widely used standard dataset for document summarization. We use standard splits and preprocess data in accordance with previous works (See et al., 2017; Liu and Lapata, 2019a; Wang and Liu, 2020).

**NYT** (Sandhaus, 2008) is another popular summarization dataset. It is collected from New York Times Annotated Corpus. We preprocess and divide this dataset according to Durrett et al. (2016).

**arXiv** and **PubMed** (Cohan et al., 2018) are two newly constructed datasets for long document summarization, which are collected from arXiv.org and PubMed.com, respectively. Xiao and Carenini (2019) created oracle labels for the two datasets. We use the same split as that of Cohan et al. (2018).

### 4.2 Models for Comparison

**NeuSum** (Zhou et al., 2018) is a neural extractive model based on seq2seq framework with attention mechanism.

**BanditSum** (Dong et al., 2018) regards sentence selection as a contextual bandit problem. Policy gradient methods are used to train the model.

**JECS** (Xu and Durrett, 2019) is a compression-based summarization model that selects sentences and compresses them by pruning a dependency tree to reduce redundancy.

**BERTSUM** (Liu and Lapta, 2019a) inserts multiple segmentation tokens into document to obtain each sentence representation. It is the first BERT-based extractive summarization model. We employ its framework as the basic document encoder of our model.

**HiBERT** (Zhang et al., 2019) modifies BERT into a hierarchical structure and design an unsupervised method to pre-train it.

**DISCOBERT** (Xu et al., 2019b) is a state-of-the-art BERT-based extractive model which encodes documents with BERT and then updates sentence representations with a graph encoder. DISCOBERT builds a document graph with only sentence units based on discourse analysis, whereas our model incorporates latent topics into a document graph and produce a heterogeneous bipartite graph.

### 4.3 Implementation Details

**Hyperparameters** For the document encoder, we use "bert-base-uncased" as our pre-trained BERT version and fine-tune it for all experiments. We also implement a non-BERT version of our model by replacing the pre-trained BERT with a Bi-GRU (Chung et al., 2014) layer and set its hidden size to 768 to compare with baseline approaches without pre-trained language models fairly. For NTM, we set topic number $K$=50. The dimension size of topic representation is set to 512. We implement GNNs with DGL (Wang et al., 2019b), and the number of GAT layer is set to 2. We set the number of attention heads to 4 for topic nodes and 6 for sentence nodes with the same hidden size of 128 to keep the dimension size of node representations unchanged. We train our model for 500 epochs with 2 NVIDIA V100 cards, and the batch size is set to 8. Except for the pre-trained BERT encoder, other parameters are randomly initialized and optimized using Adam (Kingma and Ba, 2014). $\lambda$ (Eq. 11) is set to 0.85 to balance the loss of topic modeling and sentence selection. All the hyperparameters are selected via grid search on the validation set with "Rouge-2" as metric.

**Training Strategy** We consider some empirical training strategies similar with (Cui et al., 2019) to make our model efficiently converge. Specifically, we pre-train NTM for 200 epochs with a learning rate of 1$e$-3, considering its convergence speed is much slower than that of general neural networks. In joint training, the NTM parameters are trained with a learning rate of 5$e$-4, while the learning rate of other parameters is set to 1$e$-3 because the NTM is relatively stable.

## 5 Result and Analysis

This section reports our experimental results. We evaluate our model on three criteria: 1) Whether it can achieve state-of-the-art results? 2) What benefits does the latent topic contribute to summarization? To this end, we first compare our model with state-of-the-art approaches on two widely used benchmark datasets CNN/DM and NYT. Then, we evaluate our model on two scientific paper datasets to verify whether discovering latent topics can help summarize long documents. Lastly, we present ablation and case studies for further analysis.

## 5.1 Overall Performance

Table 2 presents the Rouge F1 results of different models on CNN/DM and NYT datasets. The first section reports the Lead-3 and Oracle; the second section reports the approaches without pre-trained language models; the third section reports BERT-based models; and the last section reports our models. From the results, we make the following observations. (1) When removing pre-trained language mode, the Bi-GRU version of our model outperforms all non-BERT baseline models and obtains competitive results compared with basic BERT on both datasets. (2) Our model achieves state-of-the-art results on NYT dataset, and its performance on CNN/DM dataset is on par with DISCOBERT, which is a state-of-the-art BERT-based extractive summarization model. It needs to mention that DISCOBERT relies on external discourse analysis for modeling long-range dependencies. Our model achieves highly competitive results without external tools, which proves its inherent superiority.

| Model | CNN/DM | | | NYT | | |
|---|---|---|---|---|---|---|
| | R-1 | R-2 | R-L | R-1 | R-2 | R-L |
| Lead-3 | 40.42 | 17.62 | 36.67 | 41.80 | 22.60 | 35.00 |
| Oracle | 55.61 | 32.84 | 51.88 | 64.22 | 44.57 | 57.27 |
| NeuSum (Zhou et al., 2018) | 41.59 | 19.01 | 37.98 | – | – | – |
| BanditSum (Dong et al., 2018) | 41.50 | 18.70 | 37.60 | – | – | – |
| JECS (Xu and Durrett, 2019) | 41.70 | 18.50 | 37.90 | 45.50 | 25.30 | 38.20 |
| BERT (Zhang et al., 2019) | 41.82 | 19.48 | 38.30 | 48.38 | 29.04 | 40.53 |
| BERTSUM (Liu and Lapata, 2019) | 43.25 | 20.24 | 39.63 | – | – | – |
| HiBERT (Zhang et al., 2019) | 42.37 | 19.95 | 38.83 | 49.06 | 29.70 | 41.23 |
| DISCOBERT (Xu et al., 2019) | 43.77 | **20.85** | **40.67** | 50.00 | 30.38 | 42.70 |
| Topic-GraphSum (Bi-GRU) | 41.93 | 19.15 | 38.22 | 47.90 | 28.51 | 39.86 |
| Topic-GraphSum | **44.02** | 20.81 | 40.55 | **50.04** | **30.41** | **42.77** |

Table 2: Rouge F1 results on the test set of CNN/DM and NYT datasets. The results of comparison models are obtained from respective papers, and – represents that corresponding result is not reported.

## 5.2 Long Document Summarization

Long documents typically cover multiple topics (Xiao and Carenini, 2019). We hypothesize that our model can capture important contents of an entire document by discovering latent topics, thus enhancing the summarization performance. To verify this hypothesis, we conduct additional experiments on long-form documents. Table 3 presents the results of our model and state-of-the-art public summarization systems on arXiv and PubMed datasets. The first section includes traditional approaches and Oracle; the second and third sections include abstractive and extractive models, respectively. From Table 3, our model substantially outperforms baseline models by a large margin without pre-trained BERT, and the gaps further increase when combined with BERT. We note that discourse-aware model (Cohan et al., 2018) slightly outperforms our model on R-L of PubMed dataset; a possible reason is that it explicitly leverages the section information (e.g., introduction and conclusion) of papers, which may be strong clues in selecting summary sentences. Our model achieves state-of-the-art performance on scientific paper datasets without additional features, indicating that discovering latent topics can indeed help summarize long document, consistent with aforementioned analysis.

| Model | arXiv | | | PubMed | | |
|---|---|---|---|---|---|---|
| | R-1 | R-2 | R-L | R-1 | R-2 | R-L |
| SumBasic[*] | 29.47 | 6.95 | 26.30 | 37.15 | 11.36 | 33.43 |
| LexRank[*] | 33.85 | 10.73 | 28.99 | 39.19 | 13.89 | 34.59 |
| LSA[*] | 29.91 | 7.42 | 25.67 | 33.89 | 9.93 | 29.70 |
| Oracle[+] | 53.88 | 23.05 | 34.90 | 55.05 | 27.48 | 38.66 |
| SummaRuNNer[+] | 42.91 | 16.65 | 28.53 | 43.89 | 18.78 | 30.36 |
| Seq2seq–attentive[+] | 43.58 | 17.37 | 29.30 | 44.81 | 19.74 | 31.48 |
| Seq2seq-cancat[+] | 43.62 | 17.36 | 29.14 | 44.85 | 19.70 | 31.43 |
| Cheng & Lapata (2019)[+] | 42.24 | 15.97 | 27.88 | 43.89 | 18.53 | 30.17 |
| Attn-Seq2Seq[*] | 29.30 | 6.00 | 25.56 | 31.55 | 8.52 | 27.38 |
| Pntr-Gen-Seq2Seq[*] | 32.06 | 9.04 | 25.16 | 35.86 | 10.22 | 29.69 |
| Discourse-aware[*] | 35.80 | 11.05 | 31.80 | 38.93 | 15.37 | **35.21** |
| Topic-GraphSum (Bi-GRU) | 44.71 | 18.84 | 32.58 | 46.13 | 20.91 | 33.27 |
| Topic-GraphSum | **46.05** | **19.97** | **33.61** | **48.85** | **21.76** | 35.19 |

Table 3: Rouge F1 results on the test set of arXiv and PubMed datasets. Results with * are token from Cohan et al. (2018), and results with + are token from Xiao and Carenini (2019).

## 5.3 Ablation Study

To analyze the relative contributions of different modules in summarizing documents, we compare our full model with three ablated variants: 1) **w/o NTM**, which removes the NTM module, builds a document graph with fully connected sentence nodes, and can be regarded as performing self-attention calculation on the top of BERT; 2) **w/o GAT**, which removes the graph attention layer, directly concatenates each of sentence representation with overall topic vector $T_d$ (Eq. 3), and sends them to the sentence classifier; and 3) **LDA Version**, which replaces NTM with standard LDA and randomly initializes each topic representation.

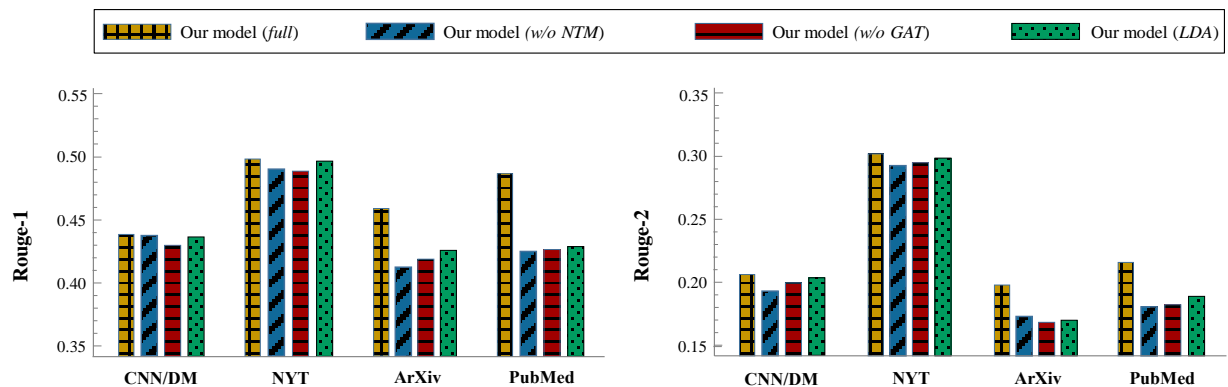

Figure 2. Rouge-1 and -2 results of our full model and three ablated variants on four datasets.

Figure 2 shows the results of different variants on four datasets, from which we can make the following observations. 1) Our full model outperforms all variants on four datasets, which proves that each module is necessary and combining them can help our model achieve the best performance. 2) When NTM module is removed or using LDA instead, the performance on arXiv and PubMed datasets declines dramatically, whereas on CNN/DM and NYT datasets, the results are competitive with our full model. A possible reason lies in that news documents are relatively short, which leads to the data sparsity problem and thus reduces the effect of topic models. 3) Similarly, when GAT is removed, the performance of scientific paper datasets has decreased more significantly than that of news datasets. This phenomenon indicates that inter-sentence relationships are especially important for summarizing long documents. 4) The LDA topic model can also boost the performance, but the gain of LDA is much fewer than that of NTM for long documents; a possible reason is that LDA and neural networks are inevitably disconnected, whereas NTM can be jointly optimized with the document encoder and graph networks, which can mutually improve each module (Wang et al., 2019).

## 5.4 Analysis of Latent Topics

In this subsection, we conduct experiments to better understand how latent topics help summarize documents. To this end, we define the topical weight of a sentence as the weighted summation of attention score between each topic and the sentence, i.e.,

**Article:** ...... For instance, one has to choose among millions of candidate commodities to shop online. Comprehensive exploration is infeasible @xcite. As a result, various recommendation approaches have been proposed to help filtering the relevant information. ...... Our present study presents a physics perspective and utilizes microscopic interactions to explain and predict macroscopic behaviors of recommender systems @xcite. In this paper, we use the Gini coefficient to measure the dispersion in item popularity @xcite. We note that a small dispersion implies similar popularity among items ......

**Golden Summary:** Recommender systems are promising ways to filter the overabundant information in modern society. their algorithms help individuals to explore decent items, but it is unclear how they allocate popularity among items. In this paper, we simulate successive recommendations and measure their influence on the dispersion of item popularity by Gini coefficient. ......

**Article:** ...... One in a million: New York alone is believed to be home to more than one million pigeons. I've also had the cops called on me on three separate occasions while I was birding on my own block. Feral pigeons find the ledges of buildings to be a substitute for sea cliffs, have become adapted to urban life, and are abundant in towns and cities throughout much of the world. ...... This was huge because it enabled me to start catching the birds in flight. I like when the pigeons dip below the horizon, this way you have the nice contrast of the crisp bird in focus with the colorful, dreamy bokeh [blur] of the backdrop. Pigeons are exceptional fliers, they are constantly on the look out for predators and must cooperate with each other to stay on top of these threats. ......

**Golden Summary:** New York alone is believed to be home to more than one million pigeons. Feral pigeons find the ledges of buildings to be a substitute for sea cliffs. Pigeons are exceptional fliers, constantly on the look out for predators. They must cooperate with each other to avoid predators.

Figure 3. Visualized results of sentence topical weight. The degree of highlighting represents the overall relevance of the sentence and all topics. Underlined sentences are model-selected summary. The left document is from PubMed dataset, and the right document is from CNN/DM dataset.

$$TW_i = \sum_{j=1}^{K} \theta^{(j)} \alpha_{j,i}, \qquad (12)$$

where $TW_i$ represents the topical weight of the $i$-th sentence. $\theta$ is the topic distribution of the document learned by NTM described in Section 3.2, and $\theta^{(j)}$ represents the weight of $j$-th topic in document. $\alpha_{j,i}$ (Eq. 5) is the attention score from the $j$-th topic node to the $i$-th sentence node.

Figure 3 shows two examples of visualized sentence topical weights. The ground-truth summary sentences have relatively high topical weights, and the final selected sentences highly overlap with these topical sentences. From such observation, we can have an intuitive understanding of how our model works. First, our model learns sentence representations and discovers latent topics, individually. Second, the graph attention layer builds semantic relationships between sentences and topics and then roughly selects important contents on the basis of topical information. Finally, our model accurately selects summary sentences by integrating all features, such as the topical relevance to the document, context information, and inter-sentence relationships. This process may explain why our model is effective for long documents. Latent topics can help our model preselect salient texts; thus, further selection can mainly focus on these fragments rather than entire document.

## 6 Conclusion and Future Work

In this paper, we systematically explore the effects of latent topics for document summarization, and propose a novel graph-based extractive summarization model, which allows joint learning of latent topics and leverages them to enrich sentence representations via a heterogeneous graph neural network. The experimental results on four well-studied datasets demonstrate that our model not only achieves results on par with state-of-the-art summarization models on news article datasets but also significantly outperforms existing approaches on scientific paper datasets, indicating its strong robustness in various document genres and lengths. Further explorations on incorporating more types of semantic units (e.g. keywords and entities) into document graph for enhancing the performance of summarization will be addressed in our future work.

**Acknowledgements**

This work is supported by grant from the National Natural Science Foundation of China (No. 61672192). We thank anonymous reviewers for their helpful comments on various aspects of this work.